\documentclass[10pt,twocolumn,letterpaper]{article}
\usepackage{cvpr}              
\usepackage{amsmath,amsfonts}
\usepackage{times}
\usepackage{algorithmic}
\usepackage{array}
\usepackage{textcomp}
\usepackage{stfloats}
\usepackage{url}
\usepackage{verbatim}
\usepackage{graphicx}
\usepackage{booktabs}
\usepackage{multirow}
\usepackage{tikz}

\definecolor{cvprblue}{rgb}{0.21,0.49,0.74}
\usepackage[pagebackref,breaklinks,colorlinks,allcolors=cvprblue]{hyperref}

\def\paperID{5742} 
\def\confName{CVPR}
\def\confYear{2026}

\title{End-to-End Motion Capture from Rigid Body Markers with Geodesic Loss}

\author{Hai Lan\\
University of Chinese Academy of Sciences\\
{\tt\small lanhai09@mails.ucas.ac.cn}
\and
Zongyan Li\\
Fujian Medical University\\
{\tt\small secondauthor@i2.org}
\and
Jianmin Hu\\
Fujian Medical University\\
{\tt\small secondauthor@i2.org}
\and
Jialing Yang\\
Fuzhou University\\
{\tt\small 248527094@fzu.edu.cn}
\and
Houde Dai\\
FJIRSM\\
Chinese Academy of Sciences\\
{\tt\small dhd@fjirsm.ac.cn}
}

\begin{document}
\maketitle
\begin{abstract}
    Marker-based optical motion capture (MoCap), while long regarded as the gold standard for accuracy, faces practical challenges, such as time-consuming preparation and marker identification ambiguity, due to its reliance on dense marker configurations, which fundamentally limit its scalability. To address this, we introduce a novel fundamental unit for MoCap, the Rigid Body Marker (RBM), which provides unambiguous 6-DoF data and drastically simplifies setup. Leveraging this new data modality, we develop a deep-learning-based regression model that directly estimates SMPL parameters under a geodesic loss. This end-to-end approach matches the performance of optimization-based methods while requiring over an order of magnitude less computation. Trained on synthesized data from the AMASS dataset, our end-to-end model achieves state-of-the-art accuracy in body pose estimation. Real-world data captured using a Vicon optical tracking system further demonstrates the practical viability of our approach. Overall, the results show that combining sparse 6-DoF RBM with a manifold-aware geodesic loss yields a practical and high-fidelity solution for real-time MoCap in graphics, virtual reality, and biomechanics.
\end{abstract}

\section{Introduction}


High-precision human motion capture (MoCap) technology plays a crucial role in computer graphics, embodied intelligence, and digital human modeling. While inertial measurement units (IMUs) and vision-based markerless MoCap systems offer low-cost and in-the-wild solutions, marker-based optical MoCap remains the de facto gold standard in clinical rehabilitation~\cite{fern2012biomechanical}, biomechanics~\cite{wishaupt2024applicability}, and film production, owing to its superior spatial accuracy and temporal resolution. This is achieved by tracking the 3D positions of reflective markers attached to the subject's skin, from which the subject's skeletal motion can then be estimated with high fidelity either through anatomical protocol-based mappings~\cite{kadaba1990measurement, wu2002isb, wu2005isb, hawk2008} or by solving an optimization problem that minimizes marker reconstruction error~\cite{amass}.

\begin{figure}[t]
    \centering
    \includegraphics[width=0.8\linewidth]{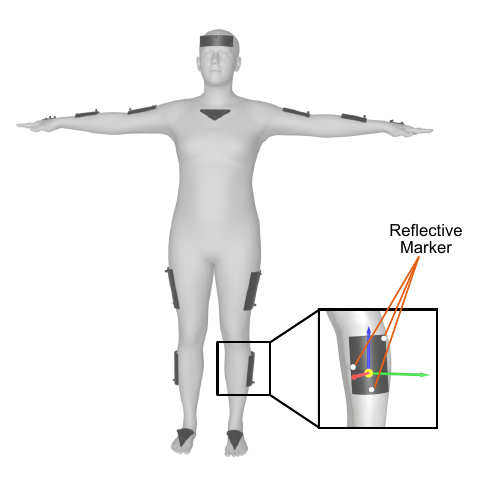}
    \caption{Schematic of the Rigid Body Marker (RBM) and its sparse placement on the human body for capturing full-body motion.}
    \label{fig:rbm}
\end{figure}

However, traditional optical MoCap typically relies on a dense set of markers distributed across the body to ensure high accuracy, which introduces several challenges. First, the preparation process is time-consuming and labor-intensive, limiting the practicality of MoCap in large-scale experiments, for example, in clinical cohort studies where multiple participants are required to perform the same standardized tasks for assessment. Second, to achieve precise measurements, markers must be affixed directly to the skin; the requirement for minimal clothing and the discomfort associated with marker removal raise ethical concerns, particularly when applied to non-professional populations. Furthermore, markers are inherently unlabeled: once a marker is occluded and reappears, it may be assigned a new index, disrupting temporal consistency and scrambling the order of the input sequence. This ambiguity impedes the estimation algorithms from reconstructing consistent and accurate poses. Although recent algorithms~\cite{SOMAICCV2021,kim2024damo} have mitigated some of these issues, their additional relabeling procedures increase implementation complexity and reduce computational efficiency. 


Embracing these challenges, we rethink the fundamental unit of marker-based MoCap and design Rigid Body Marker (RBM) to replace the traditional single-point marker. As shown in Fig.~\ref{fig:rbm}, each RBM consists of multiple reflective markers mounted on a 3D-printed rigid plate that is ergonomically shaped to fit the body contour. These modules can be fixed on the body using adjustable nylon straps, enabling fast and comfortable attachment without adhesives. Furthermore, each RBM features a unique spatial configuration: the number and placement of mounted reflective markers are individually optimized for different body parts, ensuring that the optical tracking system can unambiguously capture the full 6-DoF motion of each module. This design greatly simplifies the preparation process and effectively eliminates marker labeling ambiguity.


Moreover, we propose a geodesic-loss-based deep learning framework tailored for RBM. Our approach employs the SMPL model~\cite{smpl} and a temporal network to map RBM's 6-DoF input to SMPL parameters. The geodesic loss accurately captures rotational deviations near discontinuities, whereas directly regressing SMPL parameters using Mean Squared Error (MSE) often yields poor performance~\cite{Zhou_2019_CVPR}. Furthermore, the geodesic loss avoids the gradient explosion associated with conventional $\arccos$-based formulations, resulting in a computationally efficient and numerically stable framework. Consequently, our method achieves state-of-the-art performance using only direct SMPL parameter supervision.

We conduct extensive experiments on generated data from AMASS dataset, complemented by qualitative evaluations using a Vicon optical tracking system on real motion data, to demonstrate our method's superior precision, robustness, and applicability. Overall, this work presents a dual contribution: the RBM concept offers a new perspective on fundamental design for marker-based MoCap, while the geodesic loss establishes a geometrically-principled standard for rotation-aware loss functions in regression-based pose estimation.


\section{Related Work}

\subsection{Optical Motion Capture}

Optical MoCap systems are primarily categorized into marker-based and markerless approaches. Among these, marker-based optical MoCap remains the gold standard for kinematic and anthropometric analysis, offering sub-millimeter spatial accuracy and high temporal resolution under controlled laboratory conditions. The foundation was laid by Cappozzo et al.~\cite{cast1995}, proposed the CAST protocol using reflective markers as anatomical landmarks to establish bone-embedded coordinate systems, a methodology later adopted in the ISB recommendations~\cite{wu2002isb, wu2005isb}. Subsequent research has focused on mitigating inherent limitations such as soft tissue artifacts and landmark identification ambiguity through techniques like kinematic clustering~\cite{cappozzo2005human}, multi-body optimization~\cite{soodmand2025multibody}, and rigid cluster of markers~\cite{collins2009six}. Notably, Kaufmann et al.~\cite{kaufmann2021pose,kaufmann2023emdb} employed wireless electromagnetic tracking sensors to obtain 6-DoF marker poses, providing both position and orientation information analogous to that in~\cite{collins2009six}. These studies collectively indicate that 6-DoF markers can mitigate landmark ambiguity and achieve comparable accuracy even with a sparser configuration. Inspired by this idea, we designed the 6-DoF RBM to alleviate the dependency on anatomical expertise and time-consuming preparation required in traditional 3-DoF dense marker settings.

An alternative direction to enhance flexibility and reduce cost is markerless MoCap, which estimates 3D human pose from monocular~\cite{kanazawaHMR18}, multi-view~\cite{joo2015panoptic}, or depth cameras~\cite{shotton2011real} without physical markers. While such approaches offer unparalleled deployment flexibility, their accuracy, particularly under occlusion~\cite{kessler2019direct} or during complex motions~\cite{needham2021accuracy}, still falls short of marker-based systems for applications demanding high precision.



\subsection{Pose Estimation Methods}

Commercial marker-based MoCap systems (e.g., Vicon, OptiTrack) typically adopt proprietary solutions that are conceptually aligned with established anatomical protocols~\cite{kadaba1990measurement, wu2002isb, wu2005isb, hawk2008}. These approaches prescribe standardized marker configurations and heuristic mappings from surface motion to underlying skeletal kinematics, making them widely regarded as reference standards in quantitative motion analysis. Consequently, recent research has primarily focused on more flexible problem settings, such as cross-configuration body parameter fitting~\cite{AMASS_MoSh, amass} and resolving marker-tracking ambiguities~\cite{SOMAICCV2021, kim2024damo}.

In contrast, a variety of pose estimation algorithms have flourished in the domain of markerless MoCap. Following the paradigm of SMPLify~\cite{smpLify2016}, which estimates 3D human pose from a single image, subsequent research has leveraged deep learning models to complement optimization-based methods. These models not only extract features from images but also provide a better initialization, reducing optimization iterations and mitigating local minima~\cite{Lassner0KBBG17, selfsupervisedTungTYF17, kanazawaHMR18, RogezWS20, kolotouros2019spin,li2021hybrik, LGD2020Song}. However, directly regressing body parameters from images often leads to suboptimal accuracy. This challenge has prompted heuristic approaches that introduce proxy tasks, such as using virtual 3D landmarks as an intermediate representation~\cite{zanfir2021thundr, ma20233d}. For the regression model itself, a significant bottleneck is the discontinuity inherent in common rotation representations (e.g., axis-angle), which hinders effective network training. The introduction of the continuous 6D rotation representation~\cite{Zhou_2019_CVPR} has been pivotal, leading to its widespread adoption in state-of-the-art regression networks~\cite{kolotouros2019spin, zanfir2021thundr}. Sharing the same motivation, instead of seeking an alternative continuous representation, our work proposes a geodesic loss that provides a continuous and geometrically accurate measure of rotational discrepancy. This significantly improves regression-based network accuracy without extra transformations between representations.


\section{Method}

\begin{figure}[t]
    \centering
    \includegraphics[width=1\linewidth]{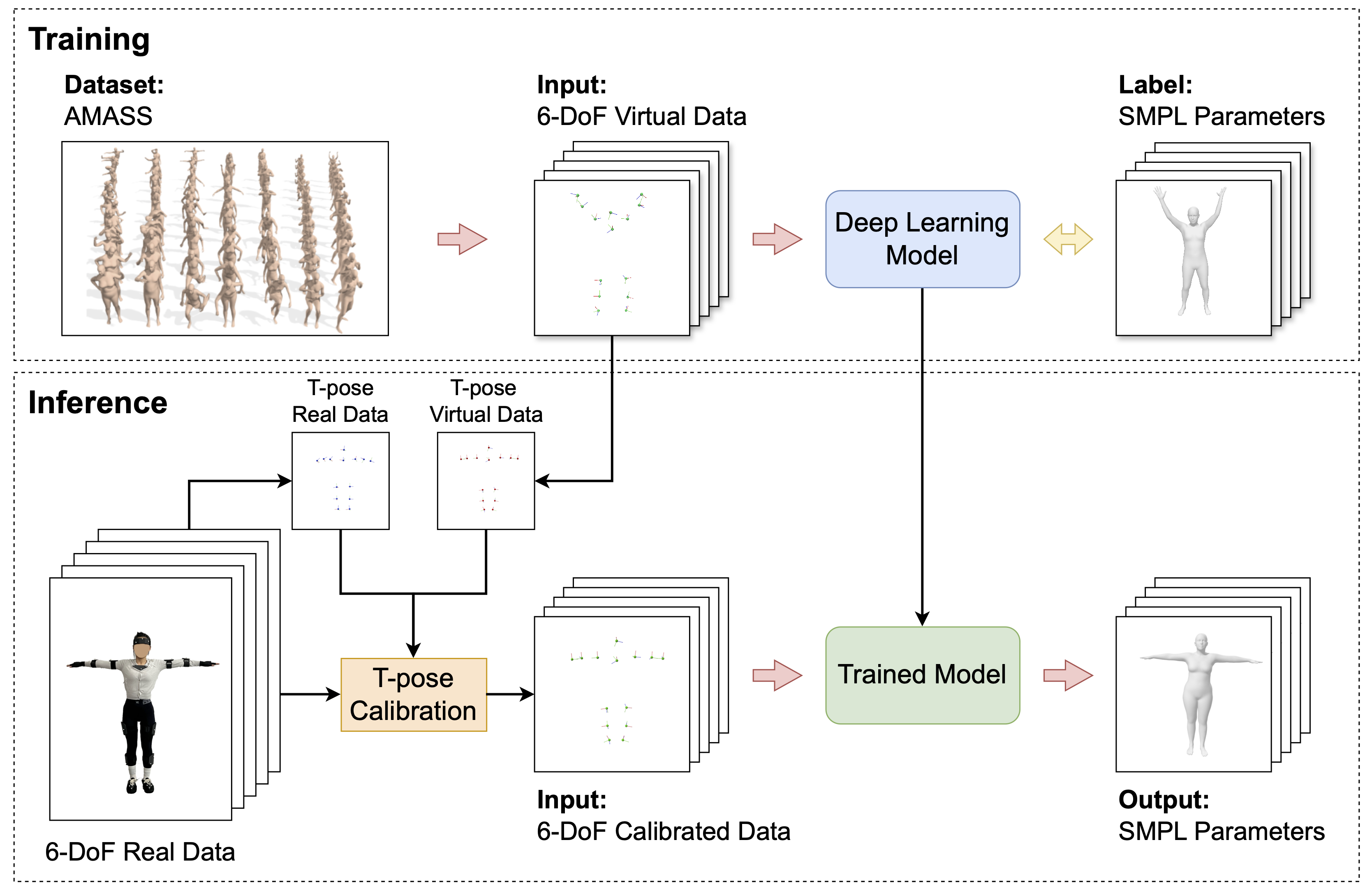}
    \caption{Pipeline of the proposed method. The upper part shows synthetic-data generation and network training; the lower part shows inference, where real inputs are calibrated and fed into the trained network to produce the final output.}
    \label{fig:infer}
\end{figure}

The overall pipeline of the proposed method is illustrated in Fig.~\ref{fig:infer}. We first synthesize virtual RBMs from the AMASS dataset to construct a training corpus. A temporal deep learning model is then trained to regress the 6-DoF inputs directly into SMPL parameters. During deployment, the 6-DoF poses of physical RBMs captured by a Vicon system are aligned to their virtual counterparts via a T-pose calibration. This alignment enables the trained model to accurately recover the human body parameters.

In the remainder of this section, we formulate the SMPL-based MoCap problem in Sec.~\ref{sec:problem}. Sec.~\ref{sec:rbm} details the RBM configuration and describes the synthesis of virtual RBMs on AMASS sequences. Secs.~\ref{sec:pose_norm} and~\ref{sec:geo_loss} present the pose normalization and the geodesic loss function, respectively. Finally, Sec.~\ref{sec:network} details the network architecture.

\subsection{Problem Statement}\label{sec:problem}

We adopt the SMPL model \cite{smpl} to parameterize human shape $\beta \in \mathbb{R}^{10}$, pose $\theta \in \mathbb{R}^{24 \times 3}$, and global translation $\gamma \in \mathbb{R}^3$. The shape parameters $\beta$ correspond to coefficients of a PCA shape-space basis, while the pose parameters $\theta$ are represented in axis-angle form, describing the 3D rotation of each of the 23 articulated joints relative to its parent in the kinematic tree, plus the root joint. The global translation $\gamma$ specifies the root position in the world frame. Given $(\beta,\theta,\gamma)$, the forward kinematics (FK) of SMPL applies a linear blend skinning function to produce a mesh $V \in \mathbb{R}^{6890 \times 3}$, from which joint locations are derived as $J_{3D} = \mathcal{J} V$ using the pre-trained regressor $\mathcal{J}$.

In a standard marker-based MoCap system, the underlying human body parameters $(\beta, \theta, \gamma)$ are recovered from the observed sensor data $\mathbf{X}$ through an inverse kinematics (IK) process. Methods like MoSh++ solve this by minimizing the discrepancy between the observed and the estimated measurements \cite{amass}. Although effective, these optimization-based methods are computationally expensive. In this work, we use a deep learning model to learn a direct mapping
\[
(\hat{\theta},\ \hat{\beta},\ \hat{\gamma}) = g(\mathbf{X})
\]
where $\mathbf{X} \in \mathbb{R}^{n\times6}$ denotes the 6-DoF measurements from $n$ RBMs. This regression-based approach enables real-time inference while maintaining high estimation accuracy. Conceptually, our work is similar to the work in~\cite{kaufmann2021pose}, which also leverages 6-DoF information from electromagnetic tracking sensors.

\subsection{RBMs and Data Synthesis}\label{sec:rbm}

We fabricate fourteen wearable RBMs via high-precision 3D printing. Dedicated fixtures allow the RBMs to be comfortably attached to the head, torso, and limbs. The 6-DoF pose data provided by the RBMs enables placement on mid-limb locations, which exhibit less soft tissue artifact than the joint landmarks required by traditional 3-DoF markers, thereby reducing a significant source of estimation error. Furthermore, each RBM integrates at least three reflective markers arranged in distinct spatial patterns, enabling the optical tracking system to unambiguously identify each module and capture its 6-DoF motion.

Acquiring a large dataset of physical measurements is a bottleneck for training deep learning models. To address this, we follow \cite{kaufmann2021pose} and synthesize virtual 6-DoF data from the AMASS dataset \cite{amass}, which offers a vast collection of SMPL parameters representing diverse human body shapes and poses. Here details the synthesis process for a single RBM; the same procedure is applied to all others.

\begin{figure}[h]
    \centering
    \includegraphics[width=\linewidth]{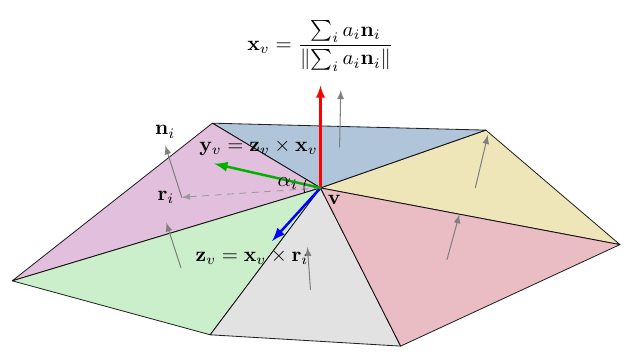}
    \caption{Illustration of the local coordinate frame construction.}
    \label{fig:lcs}
\end{figure}

The RBM's mounting location on the body mesh is defined in the local frame of its corresponding pre-selected vertex $\mathbf{v}$. As shown in Fig.~\ref{fig:lcs}, we construct the local coordinate frame $\mathcal{F}_v$ at $\mathbf{v}$ as follows:
\begin{enumerate}
    \item The \textit{x-axis} $\mathbf{x}_v$ is defined as the normalized normal vector at $\mathbf{v}$, computed via the Mean Weighted by Angle (MWA) algorithm \cite{jin2005comparison}.
    \item A \textit{reference direction} $\mathbf{r}_i$ is defined from $\mathbf{v}$ to the centroid of a pre-chosen adjacent facet. 
    \item The \textit{z-axis} $\mathbf{z}_v$ is the normalized cross product of the x-axis and the reference direction: $\mathbf{z}_v = \mathbf{x}_v \times \mathbf{r}_i$.
    \item The \textit{y-axis} $\mathbf{y}_v$ completes the right-handed orthonormal frame by the normalized cross product of the z-axis and x-axis: $\mathbf{y}_v = \mathbf{z}_v \times \mathbf{x}_v$.
\end{enumerate}
Let $\mathbf{x}_v, \mathbf{y}_v, \mathbf{z}_v \in \mathbb{R}^3$ be the basis vectors of $\mathcal{F}_v$ expressed in the world coordinate frame $\mathcal{F}_w$, and $\mathbf{p}_v$ be the world coordinates of vertex $\mathbf{v}$. The SE(3) transformation from $\mathcal{F}_v$ to $\mathcal{F}_w$ is:
\begin{equation}
\mathbf{T}_{w}^{v} = \begin{bmatrix}
    \mathbf{x}_v & \mathbf{y}_v & \mathbf{z}_v & \mathbf{p}_v \\
    0&0&0 & 1
\end{bmatrix}.
\end{equation}

The RBM's fixed offset relative to $\mathcal{F}_v$ is defined by a rotation $\mathbf{R}_o$ and a translation $\mathbf{p}_o$. The corresponding SE(3) transformation from the RBM's frame $\mathcal{F}_o$ to $\mathcal{F}_v$ is:
\begin{equation}
\mathbf{T}_{v}^{o} = \begin{bmatrix}
    \mathbf{R}_o & \mathbf{p}_o \\
    \mathbf{0} & 1
\end{bmatrix}.
\end{equation}

Given SMPL parameters $\Theta = (\beta, \theta, \gamma)$, we compute the posed body mesh vertices using FK. For each RBM, we construct the corresponding local frame $\mathcal{F}_v$, yielding $\mathbf{T}_{w}^{v}$. The global 6-DoF pose of the RBM is then obtained by concatenating the transformations:
\begin{equation}
    \mathbf{T}_{w}^{o} = \mathbf{T}_{w}^{v} \mathbf{T}_{v}^{o}.
\end{equation}

For virtual RBMs, we set the offset to $\mathbf{p}_o = [0, 0, 0.0095]^\top$ and $\mathbf{R}_o = \mathbf{I}_3$. For real-world data, the relative transform $\mathbf{T}_{v}^{o}$ is obtained through a T-pose calibration procedure.

\subsection{Pose Normalization}\label{sec:pose_norm}
The input to our deep learning model is the relative pose of the RBMs. The normalization process is detailed for position and orientation data separately.

\noindent \textbf{Position Normalization.} For the 3-DoF positional data, we first compute the centroid of all \(N\) RBM positions:
\begin{equation}
\mathbf{c} = \frac{1}{N} \sum_{i=1}^{N} \mathbf{p}_i,
\end{equation}
where $\mathbf{p}_i$ is the position of the i-th RBM. The absolute positions are then centered by subtracting this centroid, yielding relative coordinates \(\tilde{\mathbf{p}}_i = \mathbf{p}_i - \mathbf{c}\). To preserve global context, the centroid coordinate is concatenated with all relative coordinates to form the positional input feature:
\begin{equation}
\mathbf{X}_{\text{pos}} = \left[ \mathbf{c} \; \tilde{\mathbf{p}}_1 \; \tilde{\mathbf{p}}_2 \; \cdots \; \tilde{\mathbf{p}}_N \right],
\end{equation}
where \([\cdot]\) denotes the concatenation operation, resulting in \(\mathbf{X}_{\text{pos}} \in \mathbb{R}^{(N+1) \times 3}\).

\noindent \textbf{Orientation Normalization.} For the 3-DoF orientation data, we build a kinematic tree where the RBM on the chest is designated as the \textit{root} node. We keep the root node's global rotation unchanged to maintain the global orientation information. For any other node \(i\) in the kinematic tree, we compute its rotation relative to its parent node \(\text{pa}(i)\) to align with the pose parameter \(\theta\) in the SMPL model. This is achieved by multiplying the node's global rotation quaternion by the inverse of its parent's global rotation:

\begin{equation}
\tilde{\mathbf{q}}_{i} = \mathbf{q}_{\text{pa}(i)}^{*} \otimes \mathbf{q}_i,
\end{equation}
where \(\mathbf{q}_i \in \mathrm{SU}(2)\) is the global rotation quaternion of the i-th RBM, \(\mathbf{q}_{\text{pa}(i)}^{*}\) is the conjugate of the parent RBM's quaternion, and \(\otimes\) is the Hamilton product. The relative quaternion \(\tilde{\mathbf{q}}_{i}\) is then mapped to its Lie algebra \(\mathfrak{su}(2)\) to obtain a compact 3D vector:
\begin{equation}
    \tilde{\mathbf{r}}_{i} = \log(\tilde{\mathbf{q}}_{i}),
\end{equation}
where \(\log(\cdot)\) maps the quaternion $\mathbf{q}\in \mathbb{R}^4$ to an axis-angle vector \(\mathbf{r}_i \in \mathbb{R}^3\). The final orientation input feature is the collection of all rotation vectors:
\begin{equation}
\mathbf{X}_{\text{ori}} = \left[ \tilde{\mathbf{r}}_1 \; \tilde{\mathbf{r}}_2 \; \cdots \; \tilde{\mathbf{r}}_N \right],
\end{equation}
where \(\mathbf{X}_{\text{ori}} \in \mathbb{R}^{N \times 3}\).

\noindent \textbf{Final Input Vector.} The complete input to the network is formed by flattening and concatenating the position and orientation feature matrices:
\begin{equation}
\mathbf{X}_{\text{input}} = \left[\operatorname{flatten}(\mathbf{X}_{\text{pos}}) ,\ \operatorname{flatten}(\mathbf{X}_{\text{ori}})\right],
\end{equation}
where \(\mathbf{X}_{\text{input}} \in \mathbb{R}^{3+6N}\). This representation, particularly the use of relative rotations in the kinematic tree, closely mirrors the definition of SMPL's pose parameters \(\theta\), which facilitates the network's learning of the underlying mapping between RBM's orientation and joint poses.
\begin{figure*}[t]
    \centering
    \includegraphics[width=1\linewidth]{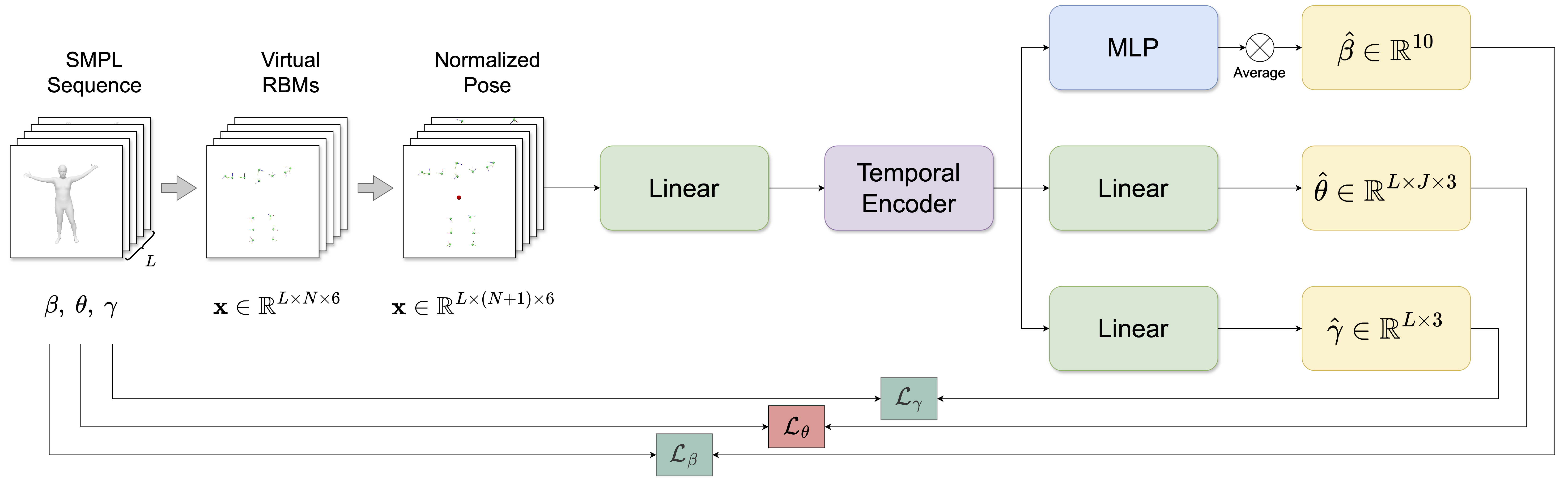}
    \caption{Overview of the network architecture and loss design.
    The network predicts SMPL parameters $(\hat{\beta}, \hat{\theta}, \hat{\gamma})$ from 6-DoF RBM input. The pose loss $\mathcal{L}_{\theta}$ employs the geodesic loss, while the shape loss $\mathcal{L}_{\beta}$ and translation loss $\mathcal{L}_{\gamma}$ use MSE loss. The total loss is computed as a weighted sum of these components.}
    \label{fig:network}
\end{figure*}
\subsection{Geodesic Loss}\label{sec:geo_loss}

In our framework, the pose parameters of the SMPL model are supervised directly in the axis-angle space. However, conventional losses such as MSE loss are sensitive to this discontinuous data. To provide a geometrically consistent and numerically stable measure of rotational discrepancy, we define a geodesic loss based on the geodesic distance on the manifold of 3D rotations \(\text{SO}(3)\). The relative rotation angle \(\Delta \theta\) between two rotations \(\mathbf{r}_1 = \alpha_1 \mathbf{n}_1\) and \(\mathbf{r}_2 = \alpha_2 \mathbf{n}_2\) is determined by the inner product of their corresponding unit quaternions \(\mathbf{q}_1\) and \(\mathbf{q}_2\) through the following relation:

\begin{equation}
    \begin{split}
        &\cos(\frac{\Delta \theta}{2}) = |\langle \mathbf{q}_1, \mathbf{q}_2\rangle | \\
        &= |\cos(\frac{\alpha_1}{2})\cos(\frac{\alpha_2}{2}) + \sin(\frac{\alpha_1}{2})\sin(\frac{\alpha_2}{2})(\mathbf{n}_1\cdot \mathbf{n}_2)|
    \end{split}
    \label{eq_inner_product_quaternion}
\end{equation}
where \(\langle \cdot, \cdot \rangle\) denotes the quaternion inner product. The canonical geodesic distance is then defined as:
\begin{equation}
    \begin{split}
        d_{\text{geo}} = \Delta \theta = 2 \arccos (|\langle \mathbf{q}_1, \mathbf{q}_2\rangle |)
    \end{split}    
\end{equation}



While \(d_{\text{geo}}\) correctly captures the shortest path between two rotations on the manifold, its gradient diverges (i.e., tends to infinity) when the two rotations are close (e.g., $|\langle \mathbf{q}_1, \mathbf{q}_2\rangle | \rightarrow 1$), leading to NaN values in the model parameters during training. To avoid this singularity and ensure stable optimization, we introduce a geodesic loss that retains the same geometric interpretation while providing smooth gradients:

\begin{equation}
    \mathcal{L}_{\text{geo}} = 4\sin^2(\frac{\Delta \theta}{2})        
\label{eq_proxy_loss}
\end{equation}

This loss avoids the singularity of the \(\arccos\) function, as its gradient is well-behaved everywhere. Furthermore, the suitability of \(\mathcal{L}_{\text{geo}}\) is twofold. First, for small angles \(\Delta \theta\), it is approximately equivalent to the squared geodesic distance $\Delta \theta^2$ due to the small-angle approximation \(\sin(x) \approx x\) for \(x \rightarrow 0\). Second, and more importantly, the function \(4\sin^2(\Delta \theta/2)\) is monotonic with respect to \(\Delta \theta\) over the interval \([0, \pi]\). This guarantees that even for large rotation errors, \(\mathcal{L}_{\text{geo}}\) still provides a gradient that consistently points in the direction of reducing the true geodesic distance \(\Delta \theta\), leading to correct and stable convergence. By substituting (\ref{eq_inner_product_quaternion}) into (\ref{eq_proxy_loss}), the final form of the loss can be computed directly from the quaternion inner product without explicitly calculating the angle \(\Delta \theta\):
\begin{equation}
    \mathcal{L}_{\text{geo}}  = 4(1- \cos^2(\frac{\Delta \theta}{2}))
        = 4(1- \langle \mathbf{q}_1, \mathbf{q}_2\rangle^2)
\end{equation}


\subsection{Network Architecture}\label{sec:network}

In contrast to recent hybrid approaches that combine iterative optimization with regression networks to mitigate the limitations of direct human body parameter regression, we demonstrate that a properly designed regression loss is sufficient. To this end, we propose a straightforward yet effective temporal regression framework. It directly maps the input sequence of 6-DoF RBM data to the output sequence of SMPL parameters, leveraging the robustness of our proposed geodesic loss. As shown in Fig.~\ref{fig:network}, the 6-DoF information of all virtual RBMs is first normalized and embedded by a linear transformation, after which the features are forwarded to temporal encoder to leverage temporal context. The encoder can be implemented with various sequence modeling backbones such as RNN, LSTM, GRU, or Transformer. The temporal features are finally mapped to the SMPL parameters, where a multilayer perceptron (MLP) head is used to regress body shape parameters $\beta$, and linear layers are used to predict pose $\theta$ and global translation $\gamma$. The model is trained using a composite loss directly supervising the ground truth of SMPL parameters without any auxiliary term:
\begin{equation}
    \mathcal{L} =  \lambda_{\beta} \mathcal{L}_{\beta} + \lambda_{\theta} \mathcal{L}_{\theta} + \lambda_{\gamma} \mathcal{L}_{\gamma}
\end{equation}
where the pose loss $\mathcal{L}_{\theta}$ is the proposed geodesic proxy loss, while the shape loss \(\mathcal{L}_{\beta}\) and the translation loss \(\mathcal{L}_{\gamma}\) use MSE loss. The coefficients \(\lambda_{\beta}\), \(\lambda_{\theta}\), and \(\lambda_{\gamma}\) are hyperparameters that balance the loss components. 

\section{Experiments}

We conduct extensive experiments on the AMASS dataset \cite{amass} to validate our framework's effectiveness and generalization. Virtual marker data is synthesized from SMPL parameters as network input. Following standard practice, we use the two largest subsets in the AMASS, CMU \cite{AMASS_CMU} (96 subjects, 1,853 sequences) and BMLrub \cite{AMASS_BMLrub} (111 subjects, 2,893 sequences). All sequences are resampled to 60 Hz and trimmed to exclude sequences shorter than two seconds (120 frames). CMU is used for training and BMLrub for evaluation, providing diverse subject and motion coverage for testing generalization.

Evaluation employs three standard metrics: Mean Per-Joint Position Error (MPJPE), its Procrustes-aligned variant (PA-MPJPE), and Mean Per-Joint Angular Error (MPJAE). All metrics are computed on the full SMPL skeleton (including root and global translation) to reflect absolute reconstruction accuracy, consistent with standard human motion estimation protocols. Our model is implemented in PyTorch and trained for 1000 epochs using the Adam optimizer with an initial learning rate of $5\times10^{-4}$. We employ a learning rate decay scheduler that multiplies the learning rate by 0.8 every 100 epochs. Unless otherwise specified, our standard architecture is composed of a two-layer Transformer with Relative Position Embedding (RPE) as the temporal encoder, with a two-layer MLP serving as the regression head for the shape parameter $\beta$. The model is trained on an NVIDIA RTX 3090 GPU.

\subsection{Quantitative Evaluation on Marker Configurations}

To evaluate the effectiveness of the proposed Rigid Body Marker (RBM) representation, we compare our approach with the conventional dense marker configuration widely adopted in optical MoCap systems. All experiments are conducted under the same network architecture and training protocol to ensure a fair comparison. The sole distinction is in the input data: the dense-marker configuration uses $N \times 3$ inputs (3D positions of $N$ markers), whereas our RBM-based configuration uses $N \times 6$ inputs (position and orientation information for $N$ RBMs).

\begin{table}[b]
    \centering
    \caption{RBM configurations used in our experiments. Each configuration defines which body segments are equipped with RBM.}
    \label{tab:rbm_config}
    \resizebox{\linewidth}{!}{

    \begin{tabular}{lccccccc}
    \toprule
    \textbf{Config} & Arm & Forearm & Hand & Thigh & Shin & Foot \\
    \midrule
    RBM-ALL & \checkmark & \checkmark & \checkmark & \checkmark & \checkmark & \checkmark \\
    RBM-A &  & \checkmark & \checkmark &  & \checkmark & \checkmark \\
    RBM-B & \checkmark &  & \checkmark & \checkmark &  & \checkmark \\
    RBM-C & \checkmark & \checkmark &  & \checkmark & \checkmark &  \\
    RBM-D &  &  & \checkmark &  &  & \checkmark \\
    RBM-E & \checkmark &  &  & \checkmark &  &  \\
    RBM-F &  & \checkmark &  &  & \checkmark &  \\
    \bottomrule
    \end{tabular}}
\end{table}

For the dense-marker baseline, we adopt the standard 53-marker configuration provided by the AMASS dataset. In contrast, our RBM configurations vary in the number of worn rigid bodies. The RBM-all setup employs all 14 rigid bodies, while RBM-A/B/C and RBM-D/E/F gradually reduce the number of rigid bodies to 10 and 6, respectively, by removing RBMs symmetrically from the limbs. The detail of the RBM configurations is shown in Table \ref{tab:rbm_config}. This design allows us to systematically assess the trade-off between marker sparsity and reconstruction accuracy.

\begin{figure}[t]
\centering
\includegraphics[width=\linewidth]{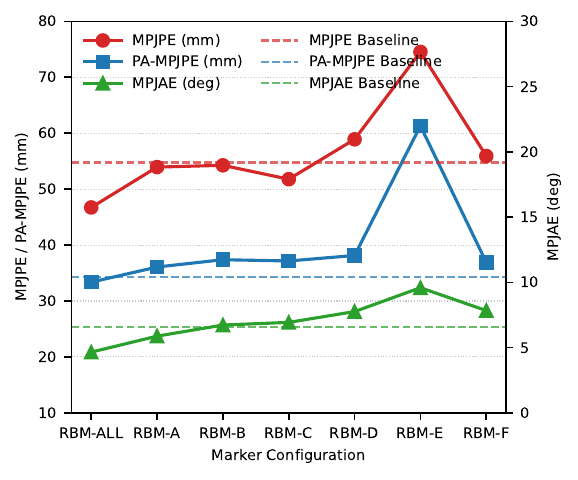}
\caption{Performance comparison of RBM configurations against the dense-marker baseline. The performance of seven RBM setups (detailed in Table \ref{tab:rbm_config}) is plotted, with the dense-marker system's result shown as a horizontal dashed line. Performance is measured by MPJPE (mm), PA-MPJPE (mm), and MPJAE (deg).}
\label{fig:exp1}
\end{figure}

Quantitative results are illustrated in Fig.~\ref{fig:exp1}. Notably, the RBM-All configuration outperforms the traditional dense marker baseline across all metrics, achieving the lowest MPJPE (46.7 mm), PA-MPJPE (33.4 mm), and MPJAE (4.7°). This optimal performance highlights the benefit of the full 6-DoF input in capturing comprehensive positional and orientational information.

Furthermore, as the number of RBMs decreases, performance degrades gradually, as expected. However, even the sparsest configurations (RBM-D and RBM-F), each containing only 6 RBMs, exhibit performance comparable to the dense marker baseline. This demonstrates the remarkable robustness and informational redundancy of the RBM-based representation.

A more nuanced asymmetry is observed when analyzing the impact of RBM placement. Eliminating distal RBMs (e.g., on hands or feet in RBM-C) results in a lower MPJPE, whereas removing proximal RBMs (e.g., on arms or thighs in RBM-A) leads to a superior MPJAE. This empirical finding suggests that distal RBMs play a more critical role in maintaining consistent joint orientation across the kinematic chain, whereas proximal RBMs are more influential for ensuring the global positional accuracy.

This phenomenon can be explained by biomechanics: proximal limb segments exhibit a more constrained range of motion relative to the torso and are generally more stable. As a result, their global pose can be effectively inferred from the orientation cues of their child segments. In contrast, distal segments undergo larger and more complex motions. This high variability makes it challenging for the regression model to learn a direct mapping from their position cues to a stable global pose. This interpretation is further corroborated by experiments with extreme sparsity, where only one RBM per limb can be retained. In this scenario, preserving the distal RBM (e.g., on the hand or feet in RBM-D) yields significantly better overall performance than preserving the proximal one (e.g., on the arm or thigh in RBM-E). Interestingly, retaining the intermediate segment RBM (e.g., on the forearm or shin in RBM-F) demonstrates a slight advantage over the most distal placement (hand or foot), a counter-intuitive insight that offers valuable empirical guidance for designing minimal yet effective RBM configurations in practice.

\subsection{Comparison with State-of-the-Art Methods}

To provide a comprehensive evaluation of our proposed method, we compare it against two state-of-the-art (SOTA) approaches based on their conceptual and structural proximity to our work:
\begin{itemize}
    \item EM-POSE \cite{kaufmann2021pose}: EM-POSE employs twelve electromagnetic tracking sensors to provide 6-DoF pose information, serving as the input for its network.
    To ensure a fair comparison, we adapt EM-POSE to our experimental setting by replacing its 12 EM sensors with our 14 RBMs, each providing full 6-DoF input. The rest of the network structure and training configuration remain unchanged.
    \item ORTH6D \cite{Zhou_2019_CVPR}: The method proposes a continuous 6D representation to alleviate the discontinuities inherent in rotation parameterizations. This representation has recently been widely adopted in SOTA markerless MoCap frameworks. Considering its conceptual similarity to our proposed geodesic proxy loss, we integrate the 6D continuous representation with MSE loss into our own temporal network architecture, while keeping all other training configurations identical. This enables a direct comparison between the two rotation modeling strategies within the same temporal modeling pipeline.
\end{itemize}

Moreover, to decouple the contribution of the temporal modeling architecture from that of the loss function, we conduct a controlled experiment by replacing the two-layer Transformer backbone with two-layer implementations of RNN, LSTM, and GRU. This setup ensures a fair comparison focused on temporal modeling efficacy, confirming the robustness of our geodesic loss across different network choices.

\begin{table}[h]
\centering
\caption{Performance comparison with state-of-the-art methods under the RBM-All configuration. FLOPs are measured for an input sequence of 120 frames.}
\resizebox{\linewidth}{!}{
\begin{tabular}{ccccc}
\toprule
\textbf{Method} & \textbf{MPJPE (mm)} & \textbf{PA-MPJPE (mm)} & \textbf{MPJAE (deg)} & \textbf{FLOPs (M)} \\
\midrule
EM-POSE & \textbf{24.729} & \textbf{18.097} & 5.325 & 3045\\
ORTH6D & 78.116 & 57.440 & 8.921 &178\\
\midrule
RNN & 51.798 & 30.354 & \textbf{4.189} &\textbf{60.6}\\
LSTM & 51.842 & 35.784 & 4.802 &164\\
GRU & 48.554 & 32.955 & 4.545 &129.5\\
Transformer & 46.725 & 33.364 & 4.652 &176.9\\
\bottomrule
\end{tabular}}
\label{tab:sota_results}
\end{table}

As shown in Table~\ref{tab:sota_results}, under the same network architecture and training protocol, our proposed geodesic loss yields a substantial improvement over the 6D continuous representation (ORTH6D). This result underscores a key insight in deep learning: when the network has sufficient representational capacity, the formulation of the loss function, encoding the problem's inductive bias, can be more critical than the choice of feature representation.

The comparison with EM-POSE, which employs a Learned Gradient Descent (LGD) framework using two networks to regress SMPL parameters and simulate an optimization loop, reveals a complementary performance profile. The optimization-based refinement of LGD yields a significantly lower MPJPE, attributable to its more accurate estimation of the global translation parameter $\gamma$. In contrast, our approach consistently achieves superior MPJAE across different backbones, indicating more precise recovery of the pose parameters $\theta$. Notably, this is accomplished with a computational cost up to 50 times lower than that of EM-POSE. This delineation suggests a promising direction for future work, integrating our efficient regression model as a high-quality initializer within optimization-based frameworks, thereby combining fast inference with the accuracy of iterative refinement.

\subsection{Ablation Study}

We further conduct an ablation study to evaluate the effectiveness of the proposed pose normalization and geodesic loss. As shown in Table~\ref{tab:ablation}, both components contribute notably to the overall performance.
Without pose normalization or geodesic loss, the model suffers a clear degradation in accuracy.
Enabling pose normalization alone slightly improves both MPJPE and MPJAE, indicating its benefit in stabilizing training across diverse motion scales.
When only the geodesic loss is applied, the model achieves a substantial reduction in MPJPE. This improvement primarily stems from more robust handling of rotational discontinuities at the root joint, which frequently occur during turning or reorientation motions and often dominate the overall position error.
The best results are obtained when both components are applied jointly, confirming their complementary benefits. Qualitative comparisons illustrating these effects are provided in the supplementary material.

\begin{table}[h]
\centering
\caption{Ablation study evaluating the impact of pose normalization and the geodesic loss under the RBM-All configuration. In the baseline setting without pose normalization, the network input comprises the RBMs’ global positions and orientations in the world coordinate frame. When replacing the geodesic loss, the baseline adopts the standard Mean Squared Error (MSE) loss on axis-angle representation.}
\resizebox{\linewidth}{!}{
\begin{tabular}{cccc}
\toprule
\textbf{Pose Norm.} & \textbf{Geo. Loss} & \textbf{MPJPE (mm)} & \textbf{MPJAE (deg)} \\
\midrule
  $\times$ & $\times$ &  143.756 & 9.356 \\
 $\surd$  & $\times$ &  126.612 &  6.034 \\
  $\times$ & $\surd$  &  67.115 &  7.227 \\
 $\surd$  & $\surd$  & \textbf{46.725} & \textbf{4.652} \\
\bottomrule
\end{tabular}}
\label{tab:ablation}
\end{table}

\subsection{Qualitative Evaluation on Real-World Data}

\begin{figure}[h]
\centering
\includegraphics[width=\linewidth]{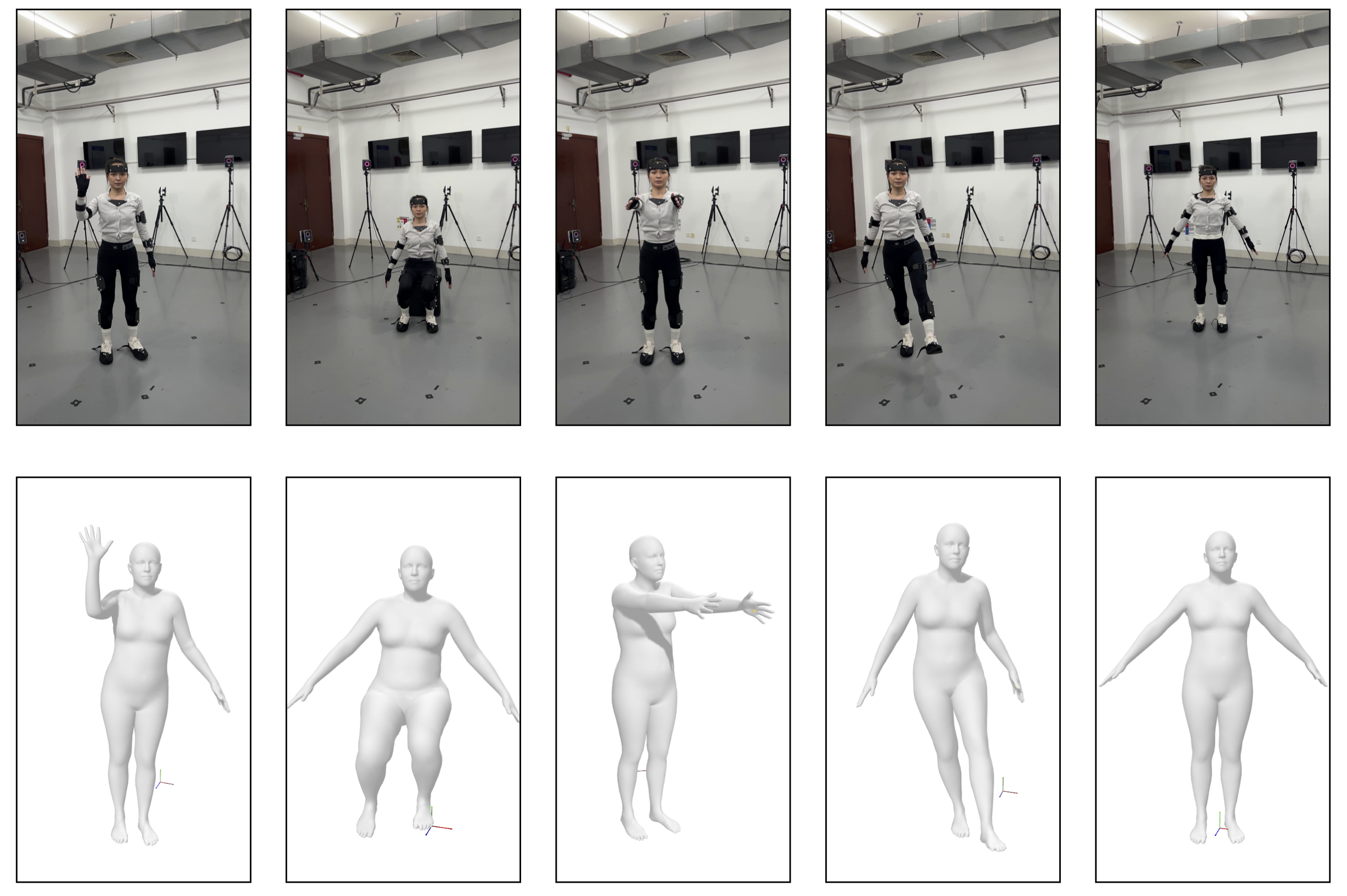}
\caption{Qualitative evaluation on real data. For five distinct actions, our method accurately reconstructs the corresponding SMPL model (bottom), closely matching the original subject (top).}
\label{fig:real_test}
\end{figure}

We present qualitative results of our method on real-world MoCap data acquired with a Vicon optical system in Fig.~\ref{fig:real_test}. All recordings were conducted with informed consent from the participants and approved by the Institutional Review Board (IRB). Additional visual comparisons are provided in the supplementary video.

\section{Limitations}

While the proposed regression-based framework demonstrates strong performance in pose estimation, several limitations remain.
First, the network exhibits limited capability in accurately regressing body shape parameters $\beta$. This is because the extra orientation information provided by the RBM provides fewer cues for inferring detailed body morphology compared to the dense positional data of a traditional marker setup.
Second, compared with optimization-based approaches that leverage a reconstruction loss, our regression-based method lacks an explicit constraint for marker fitting, resulting in reduced accuracy in global translation estimation. Moreover, the absence of smooth constraint leads to minor frame-wise discontinuities (jitter) in the reconstructed motion sequences.

\section{Conclusion}

This paper has presented a novel framework for accurate and efficient MoCap, focusing on two fundamental components of the traditional marker-based system: the physical marker and the regression loss.

First, we proposed the RBM, a minimalist hardware design that replaces traditional single-point markers with rigid modules capable of providing full 6-DoF pose data. This design significantly simplifies the preparation process by reducing the number of physical units required and eliminating marker labeling ambiguity. Experimental results show that our RBM-based system outperforms traditional dense-marker setups in accuracy while requiring far less preparation time. Comparisons among RBM configurations further yield practical guidance for optimal mounting, which is particularly valuable in large-scale applications.

Second, we developed a deep-learning-based framework using geodesic loss that directly operates on the axis-angle representation, providing a geometrically correct and numerically stable measure for rotational discrepancy. This loss enables the regression-based method to match state-of-the-art accuracy in pose estimation without any optimization refinement, reducing computational cost by orders of magnitude.

In conclusion, we demonstrate that a regression-based approach with sparse 6-DoF data, built on principled geometric design, can exceed the accuracy of dense marker systems and rival sophisticated hybrid methods, all within a simple and efficient framework. This result offers a practical and powerful tool for both marker-based and markerless MoCap algorithm development, especially in scenarios requiring high accuracy and real-time performance.

{
    \small
    \bibliographystyle{ieeenat_fullname}
    \bibliography{ref}

@conference{amass,
  title           = {{AMASS}: Archive of Motion Capture as Surface Shapes},
  author          = {Mahmood, Naureen and Ghorbani, Nima and Troje, Nikolaus F. and Pons-Moll, Gerard and Black, Michael J.},
  booktitle = {Proceedings of the IEEE/CVF International Conference on Computer Vision (ICCV)},
  pages           = {5442--5451},
  month           = oct,
  year            = {2019},
  month_numeric   = {10}
}

@article{AMASS_BMLrub,
  title           = {Decomposing Biological Motion: {A} Framework for Analysis and Synthesis of Human Gait Patterns},
  author          = {Troje, Nikolaus F.},
  year            = 2002,
  month           = sep,
  journal         = {Journal of Vision},
  volume          = 2,
  number          = 5,
  pages           = {2--2},
  doi             = {10.1167/2.5.2},
  month_numeric   = 9
}

@misc{AMASS_CMU,
  title           = {{CMU MoCap Dataset}},
  author          = {{Carnegie Mellon University}},
  url             = {http://mocap.cs.cmu.edu}
}

@article{AMASS_MoSh,
  title           = {{MoSh}: Motion and Shape Capture from Sparse Markers},
  author          = {Loper, Matthew M. and Mahmood, Naureen and Black, Michael J.},
  address         = {New York, NY, USA},
  publisher       = {ACM},
  month           = nov,
  number          = {6},
  volume          = {33},
  pages           = {220:1--220:13},
  journal         = {ACM Transactions on Graphics, (Proc. SIGGRAPH Asia)},
  url             = {http://doi.acm.org/10.1145/2661229.2661273},
  year            = {2014},
  doi             = {10.1145/2661229.2661273}
}

@article{cappozzo2005human,
  title={Human movement analysis using stereophotogrammetry: Part 1: theoretical background},
  author={Cappozzo, Aurelio and Della Croce, Ugo and Leardini, Alberto and Chiari, Lorenzo},
  journal={Gait \& posture},
  volume={21},
  number={2},
  pages={186--196},
  year={2005},
  publisher={Elsevier}
}

@article{soodmand2025multibody,
  title={Multibody kinematics optimization for motion reconstruction of the human upper extremity using potential field method},
  author={Soodmand, Iman and Herrmann, Sven and Kleist, Eric and Volpert, Annika and Wackerle, Hannes and Augat, Peter and Bader, Rainer and Woernle, Christoph and Kebbach, Maeruan},
  journal={Scientific reports},
  volume={15},
  number={1},
  pages={10411},
  year={2025},
  publisher={Nature Publishing Group UK London}
}

@inproceedings{smpl,
  title={SMPL: A skinned multi-person linear model},
  author={Loper, Matthew and Mahmood, Naureen and Romero, Javier and Pons-Moll, Gerard and Black, Michael J},
  booktitle={Seminal Graphics Papers: Pushing the Boundaries, Volume 2},
  pages={851--866},
  year={2023}
}

@inproceedings{kaufmann2021pose,
  title={Em-pose: 3d human pose estimation from sparse electromagnetic trackers},
  author={Kaufmann, Manuel and Zhao, Yi and Tang, Chengcheng and Tao, Lingling and Twigg, Christopher and Song, Jie and Wang, Robert and Hilliges, Otmar},
booktitle = {Proceedings of the IEEE/CVF International Conference on Computer Vision (ICCV)},
  pages={11510--11520},
  year={2021}
}

@inproceedings{kaufmann2023emdb,
  title={Emdb: The electromagnetic database of global 3d human pose and shape in the wild},
  author={Kaufmann, Manuel and Song, Jie and Guo, Chen and Shen, Kaiyue and Jiang, Tianjian and Tang, Chengcheng and Z{\'a}rate, Juan Jos{\'e} and Hilliges, Otmar},
  booktitle = {Proceedings of the IEEE/CVF International Conference on Computer Vision (ICCV)},
  pages={14632--14643},
  year={2023}
}

@article{needham2021accuracy,
  title={The accuracy of several pose estimation methods for 3D joint centre localisation},
  author={Needham, Laurie and Evans, Murray and Cosker, Darren P and Wade, Logan and McGuigan, Polly M and Bilzon, James L and Colyer, Steffi L},
  journal={Scientific reports},
  volume={11},
  number={1},
  pages={20673},
  year={2021},
  publisher={Nature Publishing Group UK London}
}

@article{kessler2019direct,
  title={A direct comparison of biplanar videoradiography and optical motion capture for foot and ankle kinematics},
  author={Kessler, Sarah E and Rainbow, Michael J and Lichtwark, Glen A and Cresswell, Andrew G and D'Andrea, Susan E and Konow, Nicolai and Kelly, Luke A},
  journal={Frontiers in bioengineering and biotechnology},
  volume={7},
  pages={199},
  year={2019},
  publisher={Frontiers Media SA}
}

@article{wishaupt2024applicability,
  title={The applicability of markerless motion capture for clinical gait analysis in children with cerebral palsy},
  author={Wishaupt, Koen and Schallig, Wouter and van Dorst, Marleen H and Buizer, Annemieke I and van der Krogt, Marjolein M},
  journal={Scientific reports},
  volume={14},
  number={1},
  pages={11910},
  year={2024},
  publisher={Nature Publishing Group UK London}
}

@article{collins2009six,
  title={A six degrees-of-freedom marker set for gait analysis: repeatability and comparison with a modified Helen Hayes set},
  author={Collins, Thomas D and Ghoussayni, Salim N and Ewins, David J and Kent, Jenny A},
  journal={Gait \& posture},
  volume={30},
  number={2},
  pages={173--180},
  year={2009},
  publisher={Elsevier}
}

@inproceedings{fern2012biomechanical,
  title={Biomechanical validation of upper-body and lower-body joint movements of kinect motion capture data for rehabilitation treatments},
  author={Fern'ndez-Baena, Adso and Susin, Antonio and Lligadas, Xavier},
  booktitle={2012 fourth international conference on intelligent networking and collaborative systems},
  pages={656--661},
  year={2012},
  organization={IEEE}
}

@article{jin2005comparison,
  title={A comparison of algorithms for vertex normal computation},
  author={Jin, Shuangshuang and Lewis, Robert R and West, David},
  journal={The visual computer},
  volume={21},
  number={1},
  pages={71--82},
  year={2005},
  publisher={Springer}
}

@inproceedings{Zhou_2019_CVPR,
title={On the Continuity of Rotation Representations in Neural Networks},
author={Zhou, Yi and Barnes, Connelly and Jingwan, Lu and Jimei, Yang and Hao, Li},
booktitle       = {Proceedings of the IEEE/CVF Conference on Computer Vision and Pattern Recognition (CVPR)},
month={June},
pages={5745--5753},
year={2019}
}

@inProceedings{kanazawaHMR18,
  title={End-to-end recovery of human shape and pose},
  author={Kanazawa, Angjoo and Black, Michael J and Jacobs, David W and Malik, Jitendra},
  booktitle       = {Proceedings of the IEEE/CVF Conference on Computer Vision and Pattern Recognition (CVPR)},
  pages={7122--7131},
  year={2018}
}

@inproceedings{joo2015panoptic,
  title={Panoptic studio: A massively multiview system for social motion capture},
  author={Joo, Hanbyul and Liu, Hao and Tan, Lei and Gui, Lin and Nabbe, Bart and Matthews, Iain and Kanade, Takeo and Nobuhara, Shohei and Sheikh, Yaser},
  booktitle = {Proceedings of the IEEE/CVF International Conference on Computer Vision (ICCV)},
  pages={3334--3342},
  year={2015}
}

@inproceedings{shotton2011real,
  title={Real-time human pose recognition in parts from single depth images},
  author={Shotton, Jamie and Fitzgibbon, Andrew and Cook, Mat and Sharp, Toby and Finocchio, Mark and Moore, Richard and Kipman, Alex and Blake, Andrew},
booktitle       = {Proceedings of the IEEE/CVF Conference on Computer Vision and Pattern Recognition (CVPR)},
  pages={1297--1304},
  year={2011},
  organization={Ieee}
}

@Inproceedings{kolotouros2019spin,
  Title          = {Learning to Reconstruct 3D Human Pose and Shape via Model-fitting in the Loop},
  Author         = {Kolotouros, Nikos and Pavlakos, Georgios and Black, Michael J and Daniilidis, Kostas},
  booktitle = {Proceedings of the IEEE/CVF International Conference on Computer Vision (ICCV)},

    pages={2252--2261},

  Year           = {2019}
}

@InProceedings{smpLify2016,
  title={Keep it SMPL: Automatic estimation of 3D human pose and shape from a single image},
  author={Bogo, Federica and Kanazawa, Angjoo and Lassner, Christoph and Gehler, Peter and Romero, Javier and Black, Michael J},
  booktitle={European conference on computer vision},
  pages={561--578},
  year={2016},
  organization={Springer}
}

@inproceedings{Lassner0KBBG17,
  title={Unite the people: Closing the loop between 3d and 2d human representations},
  author={Lassner, Christoph and Romero, Javier and Kiefel, Martin and Bogo, Federica and Black, Michael J and Gehler, Peter V},
  booktitle       = {Proceedings of the IEEE/CVF Conference on Computer Vision and Pattern Recognition (CVPR)},
  pages={6050--6059},
  year={2017}
}

@article{RogezWS20,
  author       = {Gr{\'{e}}gory Rogez and
                  Philippe Weinzaepfel and
                  Cordelia Schmid},
  title        = {LCR-Net++: Multi-Person 2D and 3D Pose Detection in Natural Images},
  journal      = {{IEEE} Trans. Pattern Anal. Mach. Intell.},
  volume       = {42},
  number       = {5},
  pages        = {1146--1161},
  year         = {2020},
  url          = {https://doi.org/10.1109/TPAMI.2019.2892985},
  doi          = {10.1109/TPAMI.2019.2892985},
  bibsource    = {dblp computer science bibliography, https://dblp.org}
}

@inproceedings{zanfir2021thundr,
  title={Thundr: Transformer-based 3d human reconstruction with markers},
  author={Zanfir, Mihai and Zanfir, Andrei and Bazavan, Eduard Gabriel and Freeman, William T and Sukthankar, Rahul and Sminchisescu, Cristian},
  booktitle = {Proceedings of the IEEE/CVF International Conference on Computer Vision (ICCV)},
  pages={12971--12980},
  year={2021}
}

@inproceedings{ma20233d,
  title={3D human mesh estimation from virtual markers},
  author={Ma, Xiaoxuan and Su, Jiajun and Wang, Chunyu and Zhu, Wentao and Wang, Yizhou},
  booktitle       = {Proceedings of the IEEE/CVF Conference on Computer Vision and Pattern Recognition (CVPR)},
  pages={534--543},
  year={2023}
}

@inproceedings{SOMAICCV2021,
  title = {{SOMA}: Solving Optical Marker-Based MoCap Automatically},
  author = {Ghorbani, Nima and Black, Michael J.},
  booktitle = {Proceedings of International Conference on Computer Vision (ICCV)},
  pages = {11117--11126},
  month = oct,
  year = {2021},
  doi = {},
  month_numeric = {10}
}

@inproceedings{li2021hybrik,
    title={Hybrik: A hybrid analytical-neural inverse kinematics solution for 3d human pose and shape estimation},
    author={Li, Jiefeng and Xu, Chao and Chen, Zhicun and Bian, Siyuan and Yang, Lixin and Lu, Cewu},
    booktitle       = {Proceedings of the IEEE/CVF Conference on Computer Vision and Pattern Recognition (CVPR)},
    pages={3383--3393},
    year={2021}
}

@article{cast1995,
title = {Position and orientation in space of bones during movement: anatomical frame definition and determination},
journal = {Clinical Biomechanics},
volume = {10},
number = {4},
pages = {171-178},
year = {1995},
issn = {0268-0033},
doi = {https://doi.org/10.1016/0268-0033(95)91394-T},
url = {https://www.sciencedirect.com/science/article/pii/026800339591394T},
author = {A Cappozzo and F Catani and U {Della Croce} and A Leardini},
}

@article{wu2002isb,
  title={ISB recommendation on definitions of joint coordinate system of various joints for the reporting of human joint motion—part I: ankle, hip, and spine},
  author={Wu, Ge and Siegler, Sorin and Allard, Paul and Kirtley, Chris and Leardini, Alberto and Rosenbaum, Dieter and Whittle, Mike and D D’Lima, Darryl and Cristofolini, Luca and Witte, Hartmut and others},
  journal={Journal of biomechanics},
  volume={35},
  number={4},
  pages={543--548},
  year={2002},
  publisher={Elsevier}
}

@article{wu2005isb,
  title={ISB recommendation on definitions of joint coordinate systems of various joints for the reporting of human joint motion—Part II: shoulder, elbow, wrist and hand},
  author={Wu, Ge and Van der Helm, Frans CT and Veeger, HEJ DirkJan and Makhsous, Mohsen and Van Roy, Peter and Anglin, Carolyn and Nagels, Jochem and Karduna, Andrew R and McQuade, Kevin and Wang, Xuguang and others},
  journal={Journal of biomechanics},
  volume={38},
  number={5},
  pages={981--992},
  year={2005},
  publisher={Elsevier}
}

@ARTICLE{hawk2008,
  author={Metcalf, Cheryl D. and Notley, Scott V. and Chappell, Paul H. and Burridge, Jane H. and Yule, Victoria T.},
  journal={IEEE Transactions on Biomedical Engineering}, 
  title={Validation and Application of a Computational Model for Wrist and Hand Movements Using Surface Markers}, 
  year={2008},
  volume={55},
  number={3},
  pages={1199-1210},
  doi={10.1109/TBME.2007.908087}}

@article{kadaba1990measurement,
  title={Measurement of lower extremity kinematics during level walking},
  author={Kadaba, Mrn P and Ramakrishnan, HK and Wootten, ME},
  journal={Journal of orthopaedic research},
  volume={8},
  number={3},
  pages={383--392},
  year={1990},
  publisher={Wiley Online Library}
}

@article{kim2024damo,
  title={DAMO: A Deep Solver for Arbitrary Marker Configuration in Optical Motion Capture},
  author={Kim, KyeongMin and Seo, SeungWon and Han, DongHeun and Kang, HyeongYeop},
  journal={ACM Transactions on Graphics},
  volume={44},
  number={1},
  pages={1--14},
  year={2024},
  publisher={ACM New York, NY}
}

@inproceedings{LGD2020Song,
  author       = {Jie Song and
                  Xu Chen and
                  Otmar Hilliges},
  editor       = {Andrea Vedaldi and
                  Horst Bischof and
                  Thomas Brox and
                  Jan{-}Michael Frahm},
  title        = {Human Body Model Fitting by Learned Gradient Descent},
  booktitle    = {Computer Vision - {ECCV} 2020 - 16th European Conference, Glasgow,
                  UK, August 23-28, 2020, Proceedings, Part {XX}},
  series       = {Lecture Notes in Computer Science},
  volume       = {12365},
  pages        = {744--760},
  publisher    = {Springer},
  year         = {2020},
  url          = {https://doi.org/10.1007/978-3-030-58565-5\_44},
  doi          = {10.1007/978-3-030-58565-5\_44},
  bibsource    = {dblp computer science bibliography, https://dblp.org}
}

@inproceedings{selfsupervisedTungTYF17,
  author       = {Hsiao{-}Yu Tung and
                  Hsiao{-}Wei Tung and
                  Ersin Yumer and
                  Katerina Fragkiadaki},
  editor       = {Isabelle Guyon and
                  Ulrike von Luxburg and
                  Samy Bengio and
                  Hanna M. Wallach and
                  Rob Fergus and
                  S. V. N. Vishwanathan and
                  Roman Garnett},
  title        = {Self-supervised Learning of Motion Capture},
  booktitle    = {Advances in Neural Information Processing Systems 30: Annual Conference
                  on Neural Information Processing Systems 2017, December 4-9, 2017,
                  Long Beach, CA, {USA}},
  pages        = {5236--5246},
  year         = {2017},
  url          = {https://proceedings.neurips.cc/paper/2017/hash/ab452534c5ce28c4fbb0e102d4a4fb2e-Abstract.html},
  bibsource    = {dblp computer science bibliography, https://dblp.org}
}
}

\end{document}